\documentclass[runningheads]{llncs}
\usepackage[T1]{fontenc}
\usepackage{graphicx}
\usepackage{booktabs}
\usepackage[misc]{ifsym}
\newcommand{\corr}{(\Letter)}
\usepackage{mwe}
\usepackage{xcolor}
\usepackage{graphicx}
\usepackage{subcaption}
\usepackage{amsmath}
\usepackage{amssymb}
\usepackage{amsfonts}
\usepackage{mathtools}
\usepackage{dsfont}
\usepackage{algorithm,algpseudocode}
\usepackage{multirow}
\usepackage{enumitem}
\usepackage{tikz}
\usepackage{booktabs} % for professional tables
\usepackage{subcaption} 
\usepackage{hyperref}

\DeclareMathOperator{\sgn}{sign}

\begin{document}

\title{The Simpler The Better: An Entropy-Based Importance Metric To Reduce Neural Networks' Depth}

\titlerunning{The Simpler The Better: EASIER}
% If the full title of your paper is short enough to also fit in the running head, you can omit the abbreviated paper title here. You can check as follows: if you comment out the \titlerunning line, something will appear in the header of all odd-numbered pages of your PDF from page 3 onward. This something is either the full title (in which case all is well), or the error message "Title Suppressed Due to Excessive Length". If this error message appears, you're going to want to provide an abbreviated title within the \titlerunning command, because if you won't do it, Springer will do it for you.

%N.B.: Author information (both in the \author{} and \authorrunning{} command) should only be present in the Camera-Ready Version of your paper. The version that you initially submit for review, ought to be double-blind. So, when initially submitting your paper, use:
% \author{Author information scrubbed for double-blind reviewing}
\author{Victor Qu\'etu\orcidID{0009-0004-2795-3749}\corr  \and Zhu Liao\orcidID{0009-0004-7022-5130} \and
Enzo Tartaglione\orcidID{0000-0003-4274-8298}}
% You may leave out the orcidID information, if you want to.
% Use \corr to indicate the corresponding author. Note the spacing around the \corr command. Only one author can be the corresponding author.

\authorrunning{V. Qu\'etu et al.}
%N.B.: comment out the \authorrunning{} command for the double-blind version of your paper submitted for review. Later, if your paper is accepted, use the command for the Camera-Ready Version.
% \authorrunning{A.L. Benjamin et al.}
% If there are two authors, write 'A.L. Benjamin and C.C. Broadus Jr.'
% If there are more than two authors, '[...] et al.' is used.

\institute{LTCI, T\'el\'ecom Paris, Institut Polytechnique de Paris, France \email{\{name.surname\}@telecom-paris.fr}
}

\tocauthor{Victor Qu\'etu\, Zhu Liao, Enzo Tartaglione}
\toctitle{The Simpler The Better: An Entropy-Based Importance Metric To Reduce Neural Networks' Depth}

\maketitle              % typeset the header of the contribution

\begin{abstract}

While deep neural networks are highly effective at solving complex tasks, large pre-trained models are commonly employed even to solve consistently simpler downstream tasks, which do not necessarily require a large model's complexity.
Motivated by the awareness of the ever-growing AI environmental impact, we propose an efficiency strategy that leverages prior knowledge transferred by large models. 
Simple but effective, we propose a method relying on an \textbf{E}ntropy-b\textbf{AS}ed \textbf{I}mportance m\textbf{E}t\textbf{R}ic (\textbf{EASIER}) to reduce the depth of over-parametrized deep neural networks, which alleviates their computational burden.
We assess the effectiveness of our method on traditional image classification setups. Our
code is available at \url{https://github.com/VGCQ/EASIER}.

\end{abstract}

\section{Introduction}
\label{sec:Intro}
\let\svthefootnote\thefootnote
\newcommand\freefootnote[1]{%
  \let\thefootnote\relax%
  \footnotetext{#1}%
  \let\thefootnote\svthefootnote%
}
Deep\freefootnote{This article has been accepted for publication at the European Conference on Machine Learning and Principles and Practice of Knowledge Discovery in Databases (ECML PKDD 2024).} Neural Networks (DNNs) have drastically changed the field of computer vision. They have been crucial in obtaining state-of-the-art results in several important computer vision domains, such as semantic segmentation~\cite{castillo2020auxiliary}, classification~\cite{kuang2023multi}, and object detection~\cite{yang2022rethinking}.
Beyond traditional computer vision tasks, DNNs have also impacted other fields by exhibiting unbridled potentials in natural language processing~\cite{touvron2023llama}, and multi-modal tasks~\cite{8292801}. DNNs' use is growing significantly in our lives and appears to be perennial.  

Despite DNNs have demonstrated scalability in terms of model and dataset size~\cite{hestness2017deep}, they hinder high computational demands. Indeed, neoteric architectures are made up of millions, or even billions, of parameters, resulting in billions, or even trillions, of FLoating-point OPerations (FLOPs) for a single inference~\cite{guo2022cmt}. 
Hence, these large models require enormous resources both in terms of pure hardware capacity and energy consumption, for training and deployment, which raises issues for real-time and on-device applications and also has an environmental impact. For instance,
GPT-3~\cite{brown2020language}, made of 175B parameters, emits around 200tCO$_2$eq for its training and its operational carbon footprint reached around 550tCO$_2$eq~\cite{faiz2023llmcarbon}.

The development of compression techniques, which constitute an essential means of remedying the resource-hungry nature of DNNs, has marked the research landscape over the past decade.
It is well-known that the complexity of the model is intrinsically linked to the generalizability of DNNs~\cite{hestness2017deep}, and since pre-trained architectures that can be used in downstream tasks tend to be over-parameterized, compression with no (or only slight) performance degradation is in principle possible~\cite{tartaglione2022loss}. 
To design a more efficient architecture, a set of methods has been proposed, ranging from parameter pruning~\cite{han2015learning} to the reduction of numerical precision~\cite{rastegari2016xnor}. Nonetheless, few approaches are capable of lessening the number of layers in a DNN. 
Indeed, removing single parameters or whole filters offers very few if any, practical benefits when it comes to using the model on recent computing resources, such as GPU. 
Thanks to the intrinsic parallel computation nature of GPUs or TPUs, the limitation on layer size, whether larger or smaller, comes mainly from memory caching and core availability.

In most cases, this parallelization capability avoids the need to reduce layer size, suggesting that another approach needs to be explored to address this problem.
Indeed, reducing the critical path that computations must traverse~\cite{pmlr-v202-ali-mehmeti-gopel23a} would help to relieve the DNN's computation demand, which can be achieved by strategically removing layers. Despite that existing approaches, like knowledge distillation~\cite{hinton2015distilling}, implicitly tackle this issue, the absence of performance degradation cannot be guaranteed, since a shallow target model is imposed.
This motivates the exploration of designing a method for neural networks' depth reduction while preserving optimal performance.

In this work, we present our method EASIER, which iteratively tries to reduce the depth of deep neural networks.
More precisely, EASIER identifies the average state of a given rectifier-activated neuron for the trained task. Given the definition of rectifier activation functions, EASIER can find the probability that this neuron uses one of the two regions, and hence can calculate an entropy-based metric per layer. Such a metric is then used to drive the linearization of layers toward neural network depth reduction. We summarize, here below, our key messages and contributions.
\begin{itemize}
    \item We highlight how we can potentially reduce the depth of a neural network with a marginal impact on the performance by characterizing layer degeneration (Sec.~\ref{sec:collapse}).
    \item We propose EASIER, a method relying on an entropy-based importance metric that pinpoints rectifier-activated layers that can be linearized (Sec.~\ref{sec:EASIER}).
    \item We test EASIER across multiple architectures and datasets for traditional image classification setups (Sec.~\ref{sec:results}), demonstrating that layer withdrawal can be achieved with little or no performance loss when over-parameterized networks are employed. Notably, we show the potential savings in terms of FLOPs and inference time on six different hardwares, highlighting the benefits of our method (Sec.~\ref{sec:ablation_study}).
\end{itemize}

\begin{figure}[t]
    \centering
    \includegraphics[width=0.9\columnwidth]{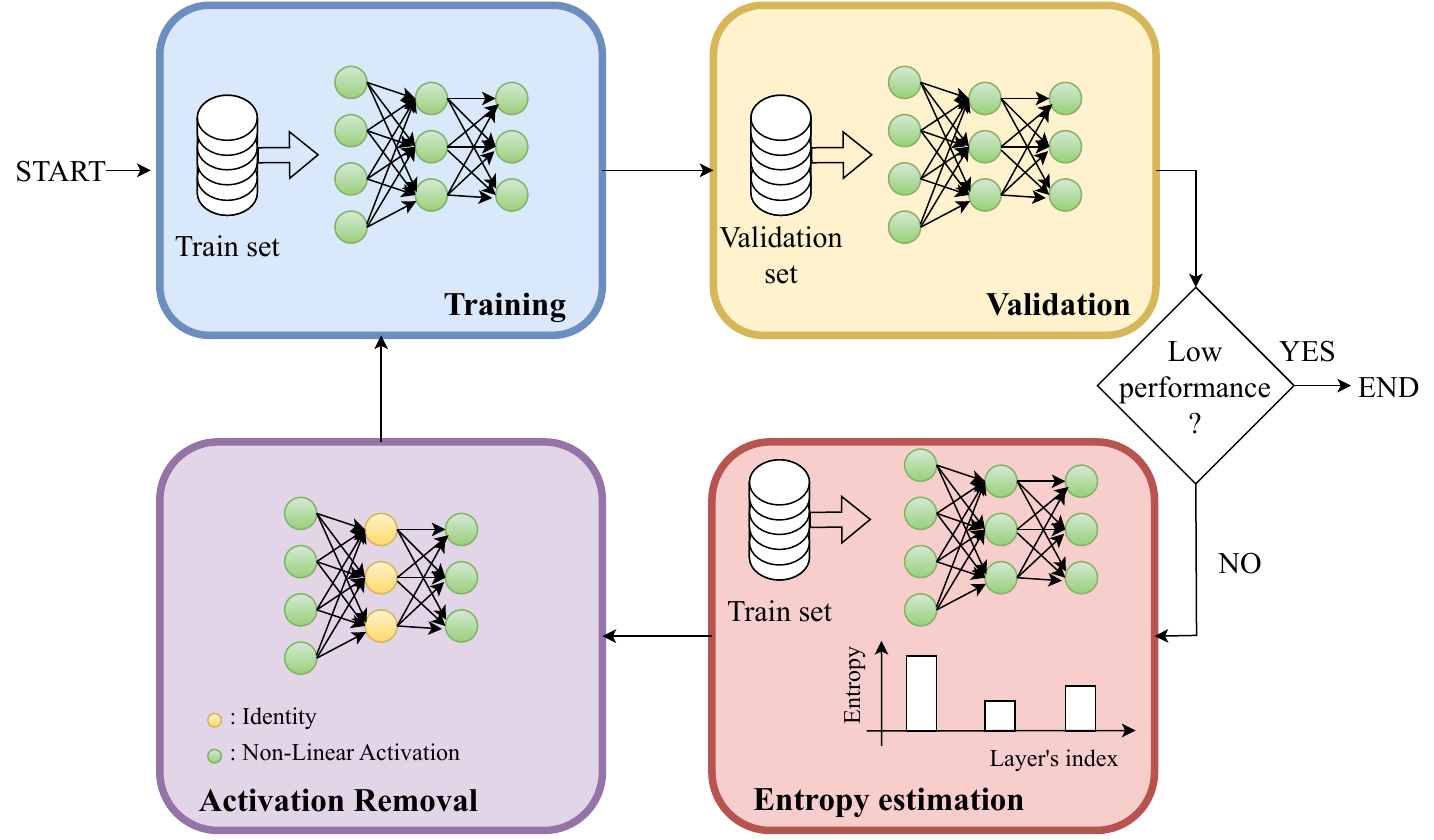}
    \caption{Overview of EASIER. We iteratively train, evaluate, and estimate the entropy on the training set and linearize the lowest-entropy layer of the neural network, until the performance drops.}
    \label{fig:Teaser_image}
\end{figure}

\section{Related Works}
\label{sec:sota}
\paragraph{Neural Architecture Search.}
Popular deep neural network architectures have mostly been designed by hand, among which we can cite VGG~\cite{simonyan2014very}, ResNet~\cite{he2016deep}, MobileNet~\cite{howard2017mobilenets} or Swin transformer~\cite{liu2021swin}.
Despite leading to remarkable performance on a variety of tasks, the design of novel architectures is time-consuming and can be prone to errors. 
Neural Architecture Search (NAS) was the answer to both these problems. 
Divided into subgroups such as evolutionary methods~\cite{mouret2015illuminating}, methods based on reinforcement learning~\cite{zoph2016neural}, and differentiable methods~\cite{liu2018darts}, NAS is finding the contemporary top-performing architectures~\cite{baymurzina2022review}.
While the firsts are based on efficient heuristic search methods based on evolution to capture global solutions of complex optimization problems~\cite{real2017large}, the second relies on goal-oriented optimization methods driven by an impact response or signal~\cite{JAAFRA201957}. Differentiable methods learn architectural paths that enable the removal of entire layers and sometimes add width to the previous ones to balance~\cite{wang2020rethinking}.
By disentangling training and searching to reduce the cost, a popular approach proposed a large once-for-all network~\cite{cai2019once} supporting diverse architectural designs. The idea was to select a sub-network within the aforementioned model without the need for additional training. 

Nonetheless, despite reducing the model size across diverse dimensions, like depth, width, kernel size, and resolution, NAS approaches, including also this work, generally need expensive computational resources to span several search space dimensions and train a super-network from scratch. In this paper, our sole focus lies on depth as the exclusive search dimension, by leveraging a pre-trained model, making easier convergence and reducing the overall training time. 

\paragraph{Neural Network Pruning.}
Neural network pruning, whose goal is to shrink a large network to a smaller one while maintaining performance by removing irrelevant weights, filters, or other structures from neural networks, has gained significant attention in neoteric works since it allows a possible model performance enhancement and an over-fitting reduction. 
On the one hand, \emph{structured} pruning focuses on removing entire neurons, filters, or channels~\cite{he2023structured,tartaglione2021serene}. On the other hand, \emph{unstructured} pruning algorithms discard weights without explicitly taking the neural network's structure into account~\cite{han2015learning,tartaglione2022loss}.
The main categories of unstructured pruning methods are magnitude-based pruning~\cite{han2015learning,louizos2018learning,h.2018to} and gradient-based pruning~\cite{lee2018snip,tartaglione2022loss}. While the first eponymous approach takes the weights' magnitude as an importance score to prune parameters, the latter uses the gradient magnitude (or its higher-order derivatives) to rank them. 
The effectiveness of these techniques was compared by~\cite{blalock2020state} and, in general, magnitude-based methods are more accurate than gradient-based.
Moreover, they are a good trade-off between complexity and competitiveness. Indeed, \cite{Gale_Magnitude}~exposed that simple magnitude pruning approaches reach similar or better results than complex methods.
From a computational perspective, in a general-purpose hardware configuration, larger benefits in terms of both memory and computation are produced by structured pruning compared to unstructured pruning, even though the reached sparsity can be significantly lower~\cite{bragagnolo2021role}.

However, a recent work~\cite{liao2023can} proposed an unstructured Entropy-Guided Pruning (EGP) algorithm, that succeeds in reducing the depth of deep neural networks by prioritizing pruning connections in low-entropy layers, leading to their entire removal while preserving performance.
Our method differs from the latter since EASIER considers a third state to calculate the entropy (Sec.~\ref{sec:Method}) and unlike EGP, our method does not involve pruning.
Although effective, EGP only allows a small number of layers to be removed. Indeed, after the removal of multiple layers, the accuracy drops dramatically. This will be verified by comparing this method with EASIER, in Sec.~\ref{sec:results}. 

\paragraph{Activation Withdrawal.}
Private inference has led to an upsurge in works on removing non-linear activations. Indeed, a high latency penalty is incurred when computing on encrypted data, which is mainly due to non-linear activations such as ReLU.
Methods such as DeepReduce~\cite{jha2021deepreduce} and SNL~\cite{cho2022selective} have been developed to reduce private inference latency. While DeepReduce includes both optimizations for ReLU dropping and knowledge distillation training to maximize the performance, the latter proposes a gradient-based algorithm that selectively linearizes ReLUs while maintaining prediction accuracy.
However, although SNL significantly reduces the number of ReLU units in the neural network, it never removes activation from an entire layer, but only from units such as pixels or channels. In contrast, our method focuses on removing activation functions at a layer level, in order to reduce the depth of deep neural networks.
Moreover, DeepReduce~\cite{jha2021deepreduce} is based on a criticality metric requiring five optimized networks per optimization iteration, resulting in the exploration of $5 \times (D-1)$ network architectures, for a network with $D$ stages, which is not very efficient at training time. On the other hand, our method does not require leveraging the knowledge of a teacher model to boost performance, saving computation at training time. 
Although left for future work, we believe our work can also be effective in accelerating private inference.

In traditional classification setups, \cite{dror2021layer} introduced Layer Folding, a technique that determines whether non-linear activations can be withdrawn, enabling the folding of adjacent linear layers into one. More specifically, PReLU activations with a trainable slope, replace ReLU-activated layers. The almost linear PReLUs are eliminated post-training, enabling the layer to be folded with its successive one.
Furthermore, a comparable channel-wise method enabling a notable reduction in non-linear units in the neural network while preserving performance was put forward by~\cite{pmlr-v202-ali-mehmeti-gopel23a}.
While the latter does not aim at reducing neural network depth, Layer Folding was originally proposed only for ReLU-activated networks. Designed for any rectifier, we will compare our method EASIER with Layer Folding and demonstrate its effectiveness in Sec.~\ref{sec:results}.

\section{Method}
\label{sec:Method}

In this section, we first highlight how we can potentially reduce the depth of a neural network with a marginal impact on performance. Based on this observation, we then derive an entropy formulation for rectifier activations, which will be at the heart of our EASIER method.

\subsection{How layers can degenerate}
\label{sec:collapse}
\begin{figure}[t]
    \centering
    \begin{subfigure}{0.49\textwidth}
        \includegraphics[width=\columnwidth]{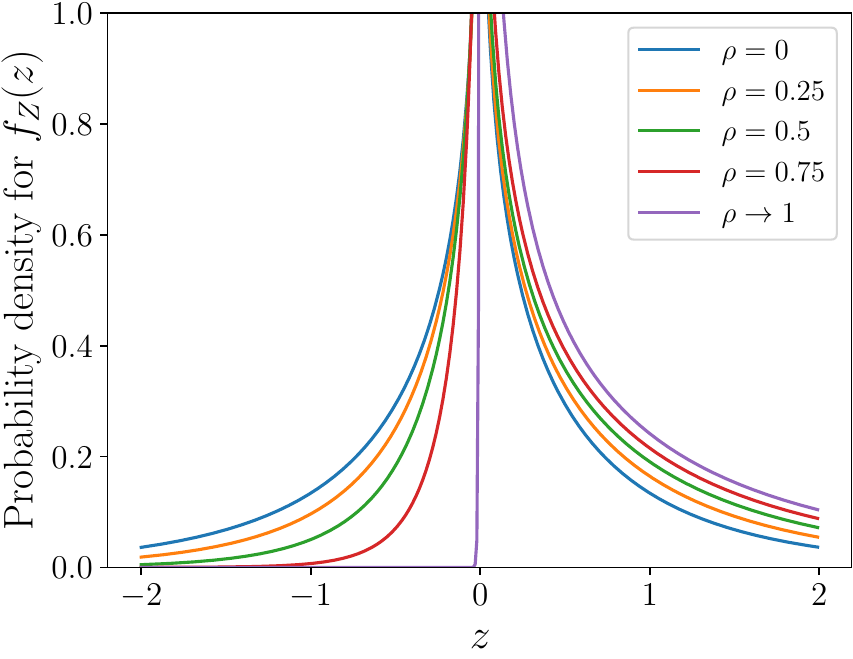}
        \caption{~}
        \label{fig:fz}
    \end{subfigure}
    \begin{subfigure}{0.49\textwidth}
        \includegraphics[width=\columnwidth]{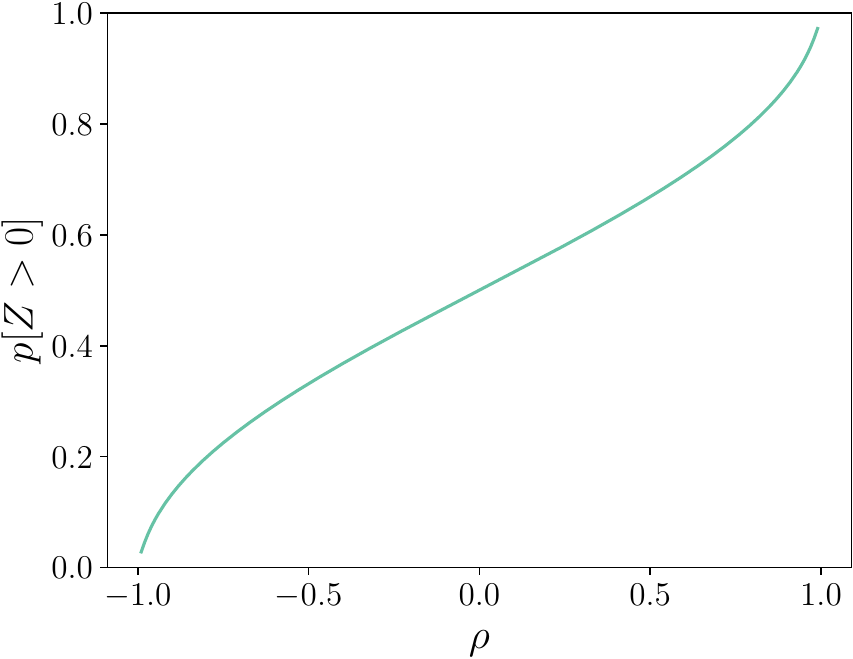}
        \caption{~}
        \label{fig:frho}
    \end{subfigure}
    \caption{Distribution of the product between $X\sim\mathcal{N}(0,1)$ and $W\sim\mathcal{N}(0,1)$ for different values of $\rho$ (a), and $p[Z>0]$ for different $\rho$ (b).}
    \label{fig:ftot}
\end{figure}

Let us define the input $\boldsymbol{x}$ for a given neuron is a sequence of random variables $~{X\sim\mathcal{N}(\mu_X, \sigma_X^2)}$. Similarly, we can assume the $N$ parameters populating such neuron, for a large $N$ limit, follow as well a Gaussian distribution, and we model it as $~{W\sim \mathcal{N}(\mu_W, \sigma^2_W)}$.
Under the assumption of $~{\mu_X=\mu_W=0}$ (for narration purposes, it is possible to derive a more general result according to~\cite{cui2016exact}), we can obtain the distribution for the pre-activation $z$ (resulting from the product of the weights and the input, modeled through the random variable $Z$), according to the result obtained by~\cite{craig1936frequency,seijas2012approach,cui2016exact}, follows the probability density function

\begin{equation} 
    f_Z(z) =\frac{1}{\pi \sigma _X\sigma _W\sqrt{1-\rho ^2}} \exp\left[\frac{\rho z}{\sigma _X\sigma _W(1-\rho ^2)}\right] K_0 \left[\frac{|z|}{\sigma _X\sigma _W(1-\rho ^2)}\right],
\end{equation} 

where $K_n$ is the n-th order modified Bessel function of the second kind and $\rho$ is the correlation coefficient between $X$ and $W$. A visual representation of its distribution is pictured in Fig.~\ref{fig:fz}. We can clearly observe the large impact of $\rho$, steering how the values will effectively be distributed. Now, let us assume the activation function of such a neuron is a rectifier function, and we are interested in observing what is the probability of the post-activation output being in the linear region: we are interested in measuring $ p[Z > 0] = 1-F_Z(0)$, where $F_Z(x)=p[Z<x]$ is the cumulative distribution function (CDF) for the density $f_Z(z)$. A visual representation of how these values are distributed for different values of $\rho$ is depicted in Fig.~\ref{fig:frho}. 

The behavior of neurons, particularly when employing rectifiers like ReLU, is tightly linked to the learning process, and $W$ becomes more and more (anti-) correlated with $X$. At $\rho\rightarrow 1$, neurons operate linearly, leading to a layer's degeneration (as the current layer becomes a linear combination with the next layer). Conversely, at $\rho\rightarrow -1$, neurons become effectively ``OFF'', leading to insignificance in their contribution. In both cases, there's a \emph{layer degeneration} that we aim to detect to reduce the neural network's depth with a marginal impact on the performance. In the next section, we will draft a metric to estimate how close a layer is to degenerating.

\subsection{Entropy for Rectifier Activations}

\label{sec:recact}
To monitor the output $y_{l,i}^{\boldsymbol{x}}$ of the $i$-th neuron from a given input $\boldsymbol{x}$ of the dataset $\mathcal{D}$, we define $\psi_l$ as the rectifier of the $l$-th layer, populated by $N_L$ neurons. Hence, by assuming that $z_{l,i}^{\boldsymbol{x}}$ is the output of the $i$-th neuron inside the $l$-th layer, we obtain:
\begin{equation}
    \label{eq:activation}
    y_{l,i}^{\boldsymbol{x}} = \psi_l(z_{l,i}^{\boldsymbol{x}}),
\end{equation}
Three possible ``states'' for the neuron can be identified from \eqref{eq:activation}:
\begin{equation}
    \label{eq:dontcare}
    s_{l, i}^{\boldsymbol{x}} = \left\{
    \begin{array}{ll}
        +1 & \text{if}\  y_{l,i}^{\boldsymbol{x}} > 0\\
        -1 & \text{if}\  y_{l,i}^{\boldsymbol{x}} < 0\\
        0 & \text{if}\  y_{l,i}^{\boldsymbol{x}} = 0.\\
    \end{array}
    \right .
\end{equation}

More precisely, for the output of the $i$-th neuron, by simply applying the $\sgn$ function to $z_{l,i}^{\boldsymbol{x}}$, we get $~{s_{l,i}^{\boldsymbol{x}} = \sgn(z_{l,i}^{\boldsymbol{x}})}$ and can hence easily pinpoint in which of these states we are.
Candidly, the neuron is in the \emph{ON state} when $s_{l,i}^{\boldsymbol{x}}=+1$, as this generally corresponds to the linear region, as opposed to the \emph{OFF state} when $s_{l,i}^{\boldsymbol{x}}=-1$ (considering that $~{\lim_{x\rightarrow -\infty}\psi(x) = 0}$).\footnote{Few exceptions to this exist, like LeakyReLU. In those occurrences, even though the activation will not converge to zero, we still choose to call it OFF state as, given the same input's magnitude, the magnitude of the output is lower.}
Since it could belong to either the ON or OFF state, the third state $s_{l, i}^{\boldsymbol{x}}=0$ is a special case, which will not be considered in the following derivation.\\
The probability (in the frequentist sense) of the i-th neuron belonging to either the ON or the OFF state can be calculated from the average over a batch of outputs for this neuron.
More precisely, we define the ON state probability as:

\begin{equation}
    \label{eq:probability}
    p(s_{l,i}\!=\!+1) = \left\{
    \begin{array}{cl}
         \displaystyle \frac{1}{S_{l,i}}\displaystyle\sum_{j=1}^{|\mathcal{D}|} s_{l,i}^{\boldsymbol{x}_j}\Theta(s_{l,i}^{\boldsymbol{x}_j}) & ~~\text{if}\ S_{l,i}\neq 0\\
        0 & ~~\text{otherwise},
    \end{array}
    \right .
\end{equation}
where
\begin{equation}
    \label{eq:counter}
    S_{l,i} = \sum_{j=1}^{|\mathcal{D}|} s_{l,i}^{\boldsymbol{x}_j}\sgn(s_{l,i}^{\boldsymbol{x}_j})
\end{equation}
is the frequency of the ON and the OFF states encountered, $|\mathcal{D}|$ is the number of input samples, and $\Theta$ is the Heaviside function.\footnote{Please be aware that additional sum and average over the entire feature map generated per input are required for convolutional layers.
}
As explained above, the third state is excluded from this count, as it can be associated with either the ON or OFF state.
We can therefore infer that since we are just concerned with the ON or OFF states, when $~{S_{l,i}\neq 0}$, $~{p(s_{l,i}\!=\!-1) = 1-p(s_{l,i}\!=\!+1)}$. 
We define as an estimator for \emph{neuron's degeneration} the entropy of the $i$-th neuron in the $l$-th layer, calculated as: 
\begin{equation}
    \label{eq:entr_neuron}
    \mathcal{H}_{l,i} =-\sum_{s_{l,i}=\pm 1} p(s_{l,i}) \log_2\left[p(s_{l,i})\right ]
\end{equation}
Given the definition in \eqref{eq:entr_neuron}, $\mathcal{H}_{l,i}=0$ can be verified in two cases:
\
\begin{itemize}%[noitemsep,nolistsep]
    \item $s_{l,i}=\!-1\ \forall j$. In this case, $z_{l,i} \leq 0\ \forall j$. The output of the $i$-th neuron is always 0 when for example employing a ReLU.
    \item $s_{l,i}=\!+1\ \forall j$. In this case, $z_{l,i} \geq 0\ \forall j$. As it belongs to the linear region, the output of the $i$-th neuron is equal to its input (or very close as in GeLU). Therefore, since there is no non-linearity between them anymore, this neuron can in principle be absorbed by the following layer. 
\end{itemize}
Please note that the case $z_{l,i} = 0\ \forall j$, can be associated with both cases, as mentioned previously, and is therefore not taken into account in the previous case disjunction.

As an estimator for \emph{layer's degeneration} we can employ the average entropy: for the $l$-th layer counting $N_l$ neuron it is
\begin{equation}
    \label{eq:entr_layer}
    \widehat{\mathcal{H}}_l = \frac{1}{N_l}\sum_i \mathcal{H}_{l,i}.
\end{equation}
We would like to have $\widehat{\mathcal{H}}_l=0$ since we target deep neural networks' depth reduction by eliminating layers with almost zero entropy. In the next section, we will present the whole framework that allows us to practically reduce the network's depth based on the layer degeneration estimator.

\subsection{EASIER}
\label{sec:EASIER}
Depicted in Alg.~\ref{alg}, we present here our method to remove the lowest-entropy layers. Indeed, the lowest-entropy layer is the one likely to make the least use of the different regions, or states, of the rectifier. Therefore, the need for a rectifier is reduced: the rectifier can be linearized entirely.
In this regard, we first train the neural network, represented by its weights at initialization $\boldsymbol{w}^{\text{init}}$, on the training set $\mathcal{D}_{\text{train}}$ (line~\ref{line:train_dense}) and evaluate it on the validation set $\mathcal{D}_{\text{val}}$ (line~\ref{line:evaluate_dense}).
As defined in \eqref{eq:entr_layer}, we then calculate the entropy $\widehat{\mathcal{H}}$ on the training set $\mathcal{D}_{\text{train}}$ for all the $L$ rectifier-activated layers, (therefore, the output layer is excluded) (line~\ref{line:evaluate_entropy}).
We then find the lowest-entropy layer (line~\ref{line:argmin}) and replace its activation with a linear one, i.e., the Identity function (line~\ref{line:linearize}). Evidently, after this step, this layer is not considered anymore. 
To recover the potential performance loss, the model is then finetuned using the same policy (line~\ref{line:retrain}) and re-evaluated on the validation set $\mathcal{D}_{\text{val}}$ (line~\ref{line:re-evaluate}).
The final model is obtained once the performance on the validation set drops below the threshold $\delta$.  %\victor{Here, maybe comment on the possibility of a one shot approach : removing 2, 4, or even 8 layers at once (Results in Appendix for CIFAR-10 ResNet-18)?}

\begin{algorithm}[t]
    \caption{Our proposed method EASIER.}
    \label{alg}
    \begin{algorithmic}[1]
        \Function{{\texttt{EASIER}}($\boldsymbol{w}^{\text{init}}$, $\mathcal{D}$, $\delta$)}{}
        \State $\boldsymbol{w}\gets $Train($\boldsymbol{w}^{\text{init}}$, $\mathcal{D}_{\text{train}}$)~\label{line:train_dense}%\alglinelabel{line:train_dense}
        \State dense\_acc $\gets $Evaluate($\boldsymbol{w}$, $\mathcal{D}_{\text{val}}$)\label{line:evaluate_dense}
        \State current\_acc $\gets$ dense\_acc
        \While{(dense\_acc - current\_acc) $>$ $\delta$} 
        \State $\widehat{\mathcal{H}} = [\widehat{\mathcal{H}}_1,\widehat{\mathcal{H}}_2, ..., \widehat{\mathcal{H}}_L]$ \algorithmiccomment{Entropy calculation on $\mathcal{D}_{\text{train}}$} \label{line:evaluate_entropy}
        \State $l$ $\gets$ argmin($\widehat{\mathcal{H}}$)\label{line:argmin}\algorithmiccomment{Finding the lowest-entropy layer}
        \State $\psi_l$ = Identity() \algorithmiccomment{Replacement of the rectifier with an Identity} \label{line:linearize}
        \State $\boldsymbol{w}\gets$ Train($\boldsymbol{w}$, $\mathcal{D}_{\text{train}}$)\algorithmiccomment{Finetune}\label{line:retrain}
        \State current\_acc $\gets$ Evaluate($\boldsymbol{w}$, $\mathcal{D}_{\text{val}}$)\label{line:re-evaluate}
        \EndWhile
        \State \textbf{return} $\boldsymbol{w}$
        \EndFunction
    \end{algorithmic}
\end{algorithm}

\begin{table}[!ht]
    \centering
    \resizebox{\columnwidth}{!}{%
    \begin{tabular}{cc|cc|cc|cc|cc}
    \toprule
        \multirow{2}{*}{\textbf{Dataset}} & \multirow{2}{*}{\textbf{Approach}} & \multicolumn{2}{c}{\textbf{ResNet-18}} & \multicolumn{2}{c}{\textbf{~Swin-T~}} & \multicolumn{2}{c}{\textbf{~MobileNetv2~}} & \multicolumn{2}{c}{\textbf{~VGG-16~}} \\
        ~ & ~ & ~~top-1~ & ~Rem.~~ & ~~top-1~ & ~Rem.~~ & ~~top-1~ & ~Rem.~~ & ~~top-1~ & ~Rem.~~ \\
    \toprule
       \multirow{4}{*}{CIFAR-10} & Dense & 92,47 & 0/17 & 91,66 & 0/12 & 93,65 & 0/35 & 93,50 & 0/15 \\ 
         & LF & 90,65 & 1/17 & 85,73 & 2/12 & 89,24 & 9/35 & 86,46 & 3/15 \\ 
         & EGP & 92,00 & 3/17 & 86,04 & 6/12 & 92,22 & 6/35 & 10,00 & 1/15 \\
         & EASIER & \bf 92,10 & \bf 8/17 & \bf 91,41 & \bf 7/12 & \bf 93,16 & \bf 12/35 & \bf 93,61 & \bf 8/15 \\
    \midrule
         \multirow{2}{*}{Tiny} & Dense & 41,26 & 0/17 & 75,78 & 0/12 & 46,54 & 0/35 & 63,94 & 0/15 \\
         \multirow{2}{*}{ImageNet} & LF & 37,86 & 4/17 & 50,54 & 1/12 & 25,88 & 12/35 & 31,44 & 6/15 \\
        \multirow{2}{*}{200} & EGP & 39,82 & 4/17 & 67,38 & 3/12 & 47,52 & 6/35 & --- & --- \\
         & EASIER & \bf 40,42 &  4/17 &  \bf 68,46 & 3/12 & \bf 48,80 & \bf 28/35 & \bf 57,60 & \bf 7/15 \\
    \midrule
       \multirow{4}{*}{PACS} & Dense & 79,70 & 0/17 & 97,30 & 0/12 & 95,50 & 0/35 & 95,40 & 0/15 \\
        ~ & LF & 82,90 & 3/17 & 87,70 & 2/12 & 79,70 & 1/35 & 93,60 & 3/15 \\
        ~ & EGP & 81,60 & 3/17 & 93,50 & 4/12 & 17,70 & 3/35 & --- & --- \\
        ~ & EASIER & \bf 84,30 & \bf 13/17 & \bf 94,30  & 4/12 & \bf 94,20 & \bf 8/35 & \bf 95,50 & \bf 4/15 \\
    \midrule
        \multirow{4}{*}{VLCS} & Dense & 68,13 & 0/17 & 83,04 & 0/12 & 81,36 & 0/35 & 82,76 & 0/15 \\
        ~ & LF & 66,91 & 5/17 & 70,92 & 1/12 & 68,87 & 2/35 & \bf 80,24 & 6/15 \\
        ~ & EGP & 70,18 & 4/17 & 78,47 & 6/12 & 45,85 & 2/35 & --- & --- \\
        ~ & EASIER & \bf 70,27 & \bf 14/17 & \bf 79,12 & 6/12 & \bf 78,56 &  \bf 4/35 & 78,84 & 6/15 \\
    \midrule
        \multirow{4}{*}{Flowers-102} & Dense & 88,88 & 0/17 & 92,70 & 0/12 & 88,50 & 0/35 & 86,47 & 0/15 \\
        ~ & LF & 77,57 & 5/17 & 63,07 & 4/12 & 2,86 & 5/35 & 87,90 & 3/15 \\
        ~ & EGP & 82,06 & 3/17 & 87,40 & 3/12 & 0,34 & 2/35 & --- & --- \\
        ~ & EASIER & \bf 83,43 & \bf 6/17 & \bf 88,89 & \bf 5/12 & \bf 88,37 & \bf 10/35 & \bf 88,32 & 3/15 \\
    \midrule
        \multirow{4}{*}{DTD} & Dense & 60,53 & 0/17 & 67,50 & 0/12 & 64,41 & 0/35 & 64,20 & 0/15 \\ 
        ~ & LF & 59,99 & 2/17 & 37,98 & 4/12 & 4,89 & 5/35 & 63,56 & 3/15 \\ 
        ~ & EGP & 59,10 & 2/17 & 60,21 & 5/12 & 2,13 & 2/35 & --- & --- \\ 
        ~ & EASIER & \bf 62,02 & \bf 3/17 & \bf 62,23 & 5/12 & \bf 63,83 & \bf 6/35 & \bf 63,62 & \bf 4/15 \\ 
    \midrule
         \multirow{4}{*}{Aircraft} & Dense & 73,36 & 0/17 & 76,39 & 0/12 & 73,36 & 0/35 & 75,85 & 0/15 \\ 
        ~ & LF & 67,60 & 2/17 & 44,76 & 4/12 & 4,98 & 4/35 & \bf 70,48 & 6/15 \\ 
        ~ & EGP & 69,04 & 2/17 & 73,27 & 5/12 & 0,99 & 2/35 & --- & --- \\ 
        ~ & EASIER & \bf 70,33 & 2/17 & \bf 74,44 & \bf 7/12 & \bf 72,55 & 4/35 & 69,70 & 6/15 \\ 
    
    \bottomrule
    \\
    \end{tabular}
    }
    \caption{Test performance (top-1) and the number of removed layers (Rem.) for all the considered setups. Dense refers to the original trained model without layer deletion. %To highlight the effectiveness of our method concerning the depth/performance tradeoff, and to enable a fair comparison with LF and EGP, EASIER results are sometimes reported with two different $\theta$.
    The best results between LF, EGP, and EASIER are in \textbf{bold}.}
    \label{tab:main_results}
\end{table}

\section{Experiments}
\label{sec:results}

In this section, we empirically evaluate the effectiveness of our proposed approach, across multiple architectures and datasets for traditional image classification setups. We compare our results with EGP~\cite{liao2023can}, an entropy-guided unstructured pruning technique, as well as the Layer Folding method~\cite{dror2021layer}.

\begin{figure}[t]
    \centering
    \begin{subfigure}{0.49\textwidth}
        \includegraphics[width=\columnwidth]{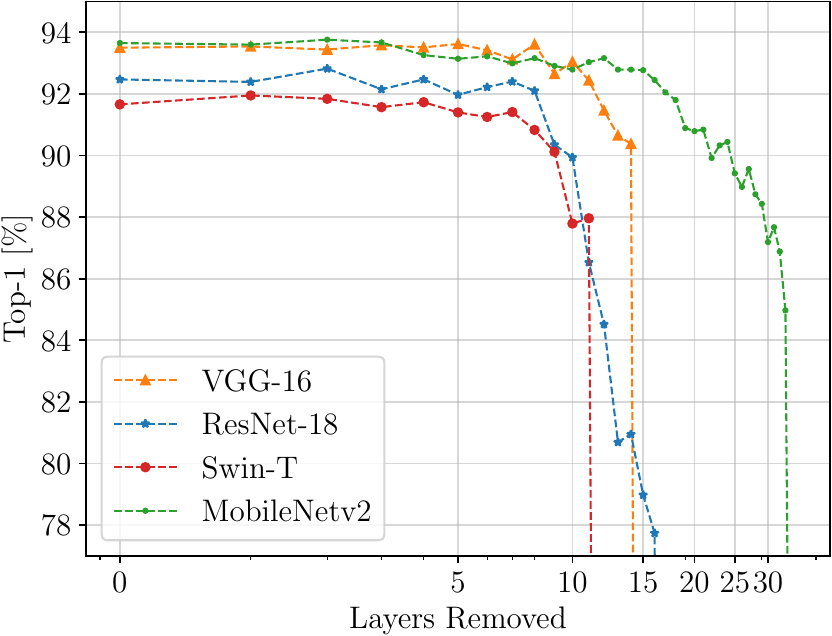}
    \caption{~}
    \label{fig:c10_top1_rem}
    \end{subfigure}
    \begin{subfigure}{0.49\textwidth}
        \centering
        \includegraphics[width=\columnwidth]{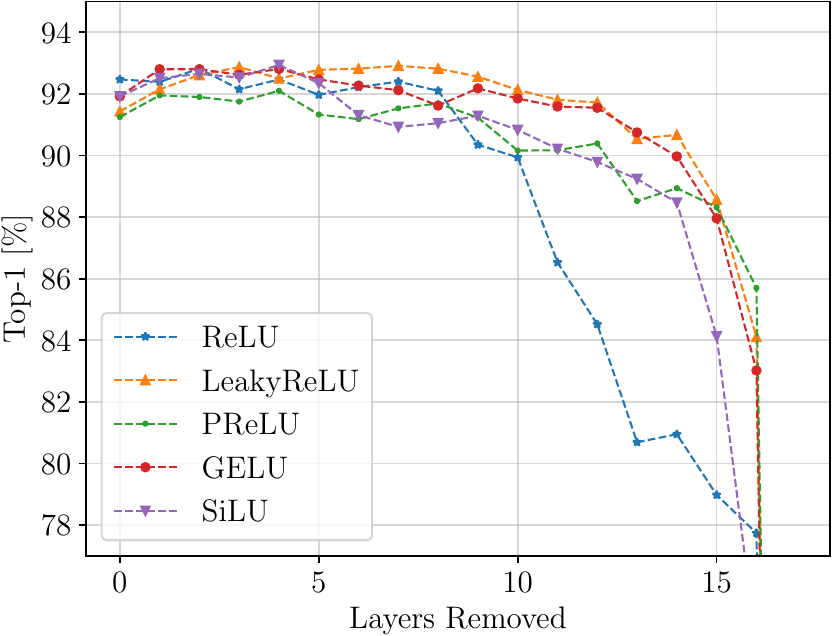}
        \caption{~}
    \label{fig:c10_top1_rem_activations}
    \end{subfigure}
    \caption{(a)~EASIER applied on ResNet-18, VGG-16, Swin-T and MobileNetv2 networks on CIFAR-10. For each model, we gradually remove non-linear layers. (b)~EASIER applied on ResNet-18 on CIFAR-10 with different rectifiers: ReLU, LeakyReLU, PReLU, GELU, and SiLU. Our method is not bound to a specific one and is effective with the most popular.}
\end{figure}

\subsection{Experimental setup}
We cover a variety of setups by evaluating our method on four popular models: ResNet-18, MobileNet-V2, Swin-T and VGG-16, trained on seven datasets: CIFAR-10~\cite{krizhevsky2009learning}, Tiny-ImageNet~\cite{le2015tiny}, PACS and VLCS from DomainBed~\cite{gulrajani2020search}, as well as Flowers-102~\cite{Nilsback08}, DTD~\cite{cimpoi14describing}, and Aircraft~\cite{maji13fine-grained}.
All the hyperparameters, augmentation strategies, and learning policies are provided in Appendix, mainly following~\cite{liao2023can} and~\cite{quetu2024dsd2}.
For ResNet-18, MobileNetv2, and VGG-16 all the ReLU-activated layers are taken into account. For Swin-T, all the GELU-activated layers are considered.
Moreover, the threshold $\delta$ is established for each dataset and architecture pair to enable a fair comparison with the existing LF and EGP approaches in terms of top-1 performance with a comparable number of removed layers.\footnote{The code and the Appendix are available at \url{https://github.com/VGCQ/EASIER}.} 

\subsection{Results}

\paragraph{A first overview.} We first test our method on a widely known dataset: CIFAR-10.
Figure~\ref{fig:c10_top1_rem} shows the test performance (Top-1) versus the number of removed layers for all the considered models on CIFAR-10, achieved with our method EASIER. Interestingly, all the models exhibit a similar depth-accuracy trend, regardless of their initial  depth. Indeed, they first all preserve their original performance, until it drops significantly once ten or so layers have been removed.

Table~\ref{tab:main_results} shows the test performance (top-1) as well as the number of removed layers (Rem.) for all the considered setups.
For each combination of dataset and architecture, the performances obtained for each iteration are shown in the tables in the Appendix.

\paragraph{Concurrent method failure in some setups.}
First, we highlight that the results for EGP on the VGG-16 architecture are not reported apart from CIFAR-10. Indeed, the EGP technique suffers from the layer collapse phenomenon~\cite{tanaka2020pruning}: by forcing a layer to have a zero-entropy, it could force it to be always in the OFF region of its activation, hence preventing the signal from passing through this layer, and therefore leading to a complete failure of the algorithm. This is what is happening on CIFAR-10, where a whole layer is pruned. Since EGP is not working with the VGG architecture on this dataset, we choose not to run the experiments for VGG on other datasets to save computations. 
Nonetheless, this is not the case with other architectures like ResNet-18, Swin-T, and MobileNetv2, which all have skip connections, leaving another alternative for the signal to pass from the input to the output, in the case the full layer is pruned. However, we also report a problem with transfer learning tasks (Flowers-102, DTD, and Aircraft) for the MobileNetv2 architecture. Indeed, from the first iteration, the EGP's pruning mechanism focuses on the last single layer before the classifier head, leading to its complete removal and hence observing the same layer collapse phenomenon given the absence of a skip/residual connection at this point.\\
Moreover, even if the results are reported in the table, we underline the failure of Layer Folding for MobileNetv2 in transfer learning setups (Flowers-102, DTD, and Aircraft). The employed auxiliary loss that encourages activations to become linear appears to have a strong effect on the final loss function. The hyperparameter balancing this regularization plays a critical role: a high value prioritizes depth reduction at a cost of performance degradation whereas a small value leads to high performance but with no layers removed. For the mentioned transfer learning task, a trade-off allowing a comparison with EASIER has not been found. This is illustrated by the results obtained on Flowers-102: even with half as many layers removed, LF achieves mediocre performance.

\paragraph{Comparison with existing approaches.} On most of the considered setups, we can observe the superiority of our method. Indeed, EASIER consistently produces models with better performance for the same number of layers removed, as observed on all the models trained on Tiny-ImageNet-200. For example, while all the methods are able to remove four layers for ResNet-18 on Tiny-ImageNet-200, EASIER achieves respectively 0,6\% and 2,56\% higher performance than EGP and LF.
Moreover, on some setups, EASIER even achieves better performance than the other competitors with more layers removed. This is the case, for example, for all the models trained on CIFAR-10. For instance, for MobileNetv2 on CIFAR-10, EASIER can remove six more layers with 0,94\% top-1 gain compared to EGP, which obtains the second-best performance in this setup. 

Nevertheless, we highlight the superiority of Layer Folding in two setups: VGG-16 trained on VLCS and Aircraft, in which the models produced by LF achieve better performance for the same number of layers removed, with performance improvements of 1,4\% and 0,78\% respectively, compared to EASIER. 

\paragraph{Comparison with the original model.}
Although on most setups (such as CIFAR-10) it succeeds in compressing models while maintaining performance similar to the original model, EASIER (but also competing methods) is not capable of compressing models without degrading performance. This is the case, for example, with Swin-T on Tiny-ImageNet-200, which displays a 7\% loss compared to the original model.
The question of a trade-off between performance and compressibility may therefore arise depending on the model's intended use.
Nevertheless, apart from VGG-16 on VLCS and Aircraft, our method produces compressed models with the closest performance to the original model compared with existing methods.

\subsection{Ablation Study}
\label{sec:ablation_study}

In this section, we first perform a study over the used rectifier, showing that our method is not bound to a specific one and is effective with any. %Table~\ref{C10_R18_rectifiers}
Fig.~\ref{fig:c10_top1_rem_activations} shows the test performance of ResNet-18 on CIFAR-10, for different rectifiers versus the number of linearized layers. Our method removes at least 8 layers with a performance improvement for GELU, LeakyReLU, and PReLU and with a marginal performance loss for ReLU and SiLU. We hypothesize that it is due to the presence of more signal in backpropagation for GELU, LeakyReLU, and PReLU. 
Moreover, to find out whether it was necessary (to maintain good performance) to train the network starting from its previous iteration weights (before a layer linearization), a randomly initialized ResNet-18 with the 8 layers selected by EASIER linearized, was re-trained on CIFAR-10 using the same learning policy. The model achieves a top-1 score of 91,56\%, down 0,54\% on the performance achieved with EASIER.
Despite being costly at training time, we concluded that to maximize the performance of the compressed model, it was important to keep training the model from its previous iteration weights.  

\begin{table}[b]
    \centering
    \begin{tabular}{ccc}
    \toprule
        Method & Top-1 & Rem. \\
    \midrule
        Dense & 92,60 & 0/17 \\ 
        EASIER $2\times$ & 92,26 & 8/17 \\ 
        EASIER $4\times$ & 92,45 & 8/17 \\ 
        EASIER $8\times$ & 91,82 & 8/17 \\ 
    \bottomrule
    \\
    \end{tabular}
    \caption{ResNet-18 on CIFAR-10.}
\label{tab:C10_R18_bottom}
\end{table}
Furthermore, to clear the way for the design of a one-shot approach, we conduct some experiments directly removing several layers at a time, for example by iteratively linearizing the 2, 4, or 8 layers with the lowest entropy. These approaches are denoted respectively EASIER $2\times$, EASIER $4\times$, and EASIER $8\times$. The results for a ResNet-18 trained on CIFAR-10 are presented in Table~\ref{tab:C10_R18_bottom}.
For fairness, we report the test performance (Top-1) for an equivalent number of layers removed (Rem.). Hence, for EASIER $2\times$ (respectively $4\times$ and $8\times$), four (respectively two and one) iterations were necessary to obtain these results. Despite removing the same number of layers, we observe that EASIER $2\times$ and EASIER $4\times$ yield similar results with a slight drop in performance compared to the original model, while EASIER $8\times$ leads to worse performance. With a performance loss of less than one percent compared to the original model, EASIER $8\times$ raises hope for the design of a one-shot approach, which would be more efficient at training time.

\begin{table}[b]
    \centering
    \resizebox{\columnwidth}{!}{%
    \begin{tabular}{c c c c c c c c}
    \toprule
        \multirow{2}{*}{\bf Rem.} &\multirow{2}{*}{\bf MFLOPs} & \multicolumn{2}{c}{\bf Inference on CPU [ms]} & \multicolumn{4}{c}{\bf Inference on GPU [ms]} \\
        ~ & ~ & Xeon E5-2640 & ~~Raspi 4~~ & Jetson Orin & ~~P2000~~ & RTX 2080  & ~~A4500~~ \\ 
    \midrule
        0/17 & 725,47 & 13,50 & 135 & 8,52 & 4,45 & 4,43 & 3,32 \\ 
        1/17 & 258,24 & 9,33 & 111 & 8,31 & 4,53 & 4,43 & 3,27 \\ 
        2/17 & 243,46 & 9,69 & 106 & 7,83 & 4,28 & 4,21 & 3,10 \\ 
        3/17 & 231,79 & 9,43 & 139 & 7,38 & 4,02 & 3,93 & 2,96 \\ 
        4/17 & 197,85 & 10,10 & 117 & 6,91 & 3,79 & 3,68 & 2,78 \\ 
        5/17 & 159,05 & 11,30 & 144 & 6,44 & 3,60 & 3,46 & 2,60 \\ 
        6/17 & 159,99 & 8,39 & 225 & 6,13 & 4,11 & 3,18 & 1,79 \\ 
        7/17 & 152,36 & 9,18 & 144 & 6,06 & 4,16 & 3,10 & 1,71 \\ 
        8/17 & 149,84 & 9,14 & 149 & 6,14 & 3,67 & 3,21 & 1,55 \\ 
    \bottomrule
    \\
    \end{tabular}
    }
    \caption{Inference time [ms] and MFLOPs of ResNet-18 on CIFAR-10.}
    \label{tab:computation}
\end{table}

Finally, Table~\ref{tab:computation} showcases the potential savings in terms of inference time and FLOPs for ResNet-18 on CIFAR-10 on six different devices, including CPUs and GPUs spanning from traditional GPU to embedded devices. In general, the fewer layers the network has, the shorter the inference time and the smaller the number of FLOPs.
However, we also observe that blindly removing layers is not sufficient to reduce computation.
Indeed, a layer removal can result in an increase in MFLOPs, as observed here at the fifth iteration, which is mainly due to the fusion of two convolutional layers, that can result in a layer having a greater size. For instance, to keep the same input/output ratio, two convolutional layers having a kernel size of 3 will fuse in a convolutional layer having a kernel size of 5.
Moreover, looking at inference times, every device shows a different trend. While larger devices, like RTX A4500, show a monotonically decreasing inference time,  for smaller devices, like P2000 or Jetson Orin, this is not always the case. We also note the same problem on CPUs, like Raspberry Pi 4, where caching is the major problem when dealing with larger kernels.

\subsection{Limitations and Future Work}

Despite being a successful approach to alleviating deep neural networks' depth, EASIER also presents some limits, which we discuss below.
\vspace{-2mm}
\paragraph{Training efficiency. } The iterative nature of our method inevitably leads to a longer training time and more intensive computations to achieve the compressed models, compared for instance, to the Layer Folding approach.
However, the increased computational cost of training can be offset by the benefits of using these models for inference. 
Indeed, since a neural network is going to be used multiple times for inference, it is also important to lessen its computational burden related to this use. As opposed to unstructured pruning which offers very few, if any, practical benefits when it comes to deploying the model in a resource-constrained system, our method reduces the critical path forward propagation undergoes, making it useful for processing on parallel systems like GPUs or TPUs, as the computational demands at inference time are reduced.\\
Nonetheless, even though the method has been thought out iteratively, there is hope for the design of a one-shot approach, which would be more efficient at training time, as shown by the results discussed in the previous ablation study.
Another way to address this problem can be to include the entropy in the minimized objective function. However, this approach is not immediately feasible as it is a non-differentiable metric. Therefore, the exploration of differentiable proxies for the layer's entropy is left as future work.
\vspace{-2mm}
\paragraph{Performance degradation.} It is difficult to compress existing parameter-efficient architectures that are not overfitting, and EASIER cannot decrease the depth of an already underfitting architecture without compromising performance, like for example Swin-T on Tiny-ImageNet-200.\\
Nevertheless, EASIER was able to demonstrate its superiority over existing methods on all the setups considered. Indeed, for the same number of removed layers, EASIER achieves the best performance or can compress more than existing approaches while maintaining performance.
We therefore believe that EASIER is a serious candidate to be considered to achieve this kind of goal.

\section{Conclusion}
\label{sec:Conclusion}

In this work, we have presented EASIER, an entropy-based method for layer withdrawal in rectifier-activated deep neural networks. 
An entropy-based importance metric has been designed to select layers to remove from the network, aiming at depth reduction while preserving high performance in the considered tasks. 
The capability and effectiveness of reducing the number of layers in a model of EASIER have been demonstrated by experiments conducted on four popular architectures across seven datasets for image classification. 
Concerned by the ever-growing AI environmental impact, we hope this work can inspire future optimizations and new ways of thinking about network design.
\vspace{-2mm}
\begin{credits}
\subsubsection{\ackname} 
This project has received funding from the European Union’s Horizon  Europe  research  and  innovation  programme  under  grant  agreement  101120237 (ELIAS).  Also, this research was partially funded by Hi!PARIS Center on Data Analytics and Artificial Intelligence.
This project was provided with computer and storage resources by GENCI at IDRIS thanks to the grant 2023-AD011013930R1 on the supercomputer Jean Zay's the V100 partition.
\end{credits}

\newpage
\appendix
\section{Details on the learning strategies employed}

The training hyperparameters used in the experiments are presented in Table~\ref{tab:learning_strategies}. Our code is available at \url{https://github.com/VGCQ/EASIER}.

\begin{table*}[!h]
    \centering
    \resizebox{\columnwidth}{!}{%
    \begin{tabular}{c c c c c c c c c c}
        \toprule
        \bf Model & \bf Dataset & \bf Epochs & \bf Batch & \bf Opt. & \bf Mom. & \bf LR & \bf Milestones & \bf Drop Factor & \bf Weight Decay \\
        \midrule
        %% CIFAR-10
         ResNet-18 & CIFAR-10 & 160 & 128 & SGD & 0.9 & 0.1 & [80, 120] & 0.1 & 1e-4 \\
         Swin-T & CIFAR-10 & 160 & 128 & SGD & 0.9 & 0.001 & [80, 120] & 0.1 & 1e-4\\
         MobileNetv2 & CIFAR-10 & 160 & 128 & SGD & 0.9 & 0.1 & [80, 120] & 0.1 & 1e-4\\
         VGG-16 & CIFAR-10 & 160 & 128 & SGD & 0.9 & 0.01 & [80, 120] & 0.1 & 1e-4\\
         \midrule
         %% Tiny ImageNet
         ResNet-18 & Tiny-ImageNet-200 & 160 & 128 & SGD & 0.9 & 0.1 & [80, 120] & 0.1 & 1e-4 \\
         Swin-T & Tiny-ImageNet-200 & 160 & 128 & SGD & 0.9 & 0.001 & [80, 120] & 0.1 & 1e-4 \\
         MobileNetv2 & Tiny-ImageNet-200 & 160 & 128 & SGD & 0.9 & 0.1 & [80, 120] & 0.1 & 1e-4 \\
         VGG-16 & Tiny-ImageNet-200 & 160 & 128 & SGD & 0.9 & 0.01 & [80, 120] & 0.1 & 1e-4 \\
         \midrule
         %% PACS
         ResNet-18 & PACS & 30 & 16 & SGD & 0.9 & 0.001 & [24] & 0.1 & 5e-4 \\
         Swin-T & PACS & 30 & 16 & SGD & 0.9 & 0.001 & [24] & 0.1 & 5e-4 \\
         MobileNetv2 & PACS & 30 & 16 & SGD & 0.9 & 0.001 & [24] & 0.1 & 5e-4 \\
         VGG-16 & PACS & 30 & 16 & SGD & 0.9 & 0.001 & [24] & 0.1 & 5e-4 \\
         \midrule
         %% VLCS
         ResNet-18 & VLCS & 30 & 16 & SGD & 0.9 & 0.001 & [24] & 0.1 & 5e-4 \\
         Swin-T & VLCS & 30 & 16 & SGD & 0.9 & 0.001 & [24] & 0.1 & 5e-4 \\
         MobileNetv2 & VLCS & 30 & 16 & SGD & 0.9 & 0.001 & [24] & 0.1 & 5e-4 \\
         VGG-16 & VLCS & 30 & 16 & SGD & 0.9 & 0.001 & [24] & 0.1 & 5e-4 \\
         \midrule
         %% Flowers-102
         ResNet-18 & Flowers-102 & 50 & 16 & Adam &  & 1e-4 & ~ & ~ & 0  \\
         Swin-T & Flowers-102 & 50 & 16 & Adam &  & 1e-4 & ~ & ~ & 0 \\
         MobileNetv2 & Flowers-102 & 50 & 16 & Adam &  & 1e-4 & ~ & ~ & 0 \\
         VGG-16 & Flowers-102 & 50 & 16 & Adam &  & 1e-4 & ~ & ~ & 0 \\
         \midrule
         %% DTD
         ResNet-18 & DTD & 50 & 16 & Adam &  & 1e-4 & ~ & ~ & 0  \\
         Swin-T & DTD & 50 & 16 & Adam &  & 1e-4 & ~ & ~ & 0 \\
         MobileNetv2 & DTD & 50 & 16 & Adam &  & 1e-4 & ~ & ~ & 0 \\
         VGG-16 & DTD & 50 & 16 & Adam &  & 1e-4 & ~ & ~ & 0 \\
         \midrule
         %% Aircraft
         ResNet-18 & Aircraft & 50 & 16 & Adam &  & 1e-4 & ~ & ~ & 0  \\
         Swin-T & Aircraft & 50 & 16 & Adam &  & 1e-4 & ~ & ~ & 0 \\
         MobileNetv2 & Aircraft & 50 & 16 & Adam &  & 1e-4 & ~ & ~ & 0 \\
         VGG-16 & Aircraft & 50 & 16 & Adam &  & 1e-4 & ~ & ~ & 0 \\
         %% ImageNet
         %ResNet-18 & ImageNet & 90 & 128 & SGD & 0.9 & 0.1 & [30, 90] & 0.1 & 1e-4 \\
          \bottomrule
          \\
    \end{tabular}
    }
    \caption{The different employed learning strategies.}
    \label{tab:learning_strategies}
\end{table*}

CIFAR-10 is augmented with per-channel normalization, random horizontal flipping, and random shifting by up to four pixels in any direction.
For the datasets of DomainBed, the images are augmented with per-channel normalization, random horizontal flipping, random cropping, and resizing to 224. The brightness, contrast, saturation, and hue are also randomly affected with a factor fixed to 0.4.
Tiny-ImageNet-200 is augmented with per-channel normalization and random horizontal flipping.
Moreover, the images of Flowers-102 are augmented with per-channel normalization, random horizontal and vertical flipping combined with a random rotation, and cropped to 224. DTD and Aircraft are augmented with random horizontal and vertical flipping, and with per-channel normalization.

Following~\cite{liao2023can} and~\cite{quetu2024dsd2}, on CIFAR-10 and Tiny-ImageNet-200, all the models are trained for 160 epochs, optimized with SGD, having momentum 0.9, batch size 128, and weight decay 1e-4. The learning rate is decayed by a factor of 0.1 at milestones 80 and 120. The initial learning rate ranges from 0.1 for ResNet-18 and MobileNetv2, 0.01 for VGG-16 to 1e-3 for Swin-T.
Moreover, on PACS and VLCS, all the models are trained for 30 epochs, optimized with SGD, having momentum 0.9, a learning rate of 1e-3 decayed by a factor 0.1 at milestone 24, batch size 16, and weight decay 5e-4.
Furthermore, on Aircraft, DTD, and Flowers-102, all the models are trained following a transfer learning strategy. Indeed, each model is initialized with their pre-trained weights on ImageNet, trained for 50 epochs, optimized with Adam, having a learning rate 1e-4 and batch size 16.

For ResNet-18, MobileNetv2, and VGG-16 all the ReLU-activated layers are taken into account. For Swin-T, all the GELU-activated layers are considered.

The results for Layer Folding are obtained using the same aforementioned training policies, with the hyper-parameters declared in~\cite{dror2021layer}. Even if the method to find a good set of hyperparameters is not provided by the authors, for the datasets on which Layer Folding was not evaluated in the original work, we tried our best to select a good set of hyperparameters. Moreover, to allow a fair comparison with our method regarding the number of layers removed, we did not use their thresholding policy to select the number of removable layers. Instead, we have chosen to remove the top-k PReLU-activated layers having the most linear slope. 

\section{Detailed Results}

For each architecture, the test performance (top-1) and the number of removed layers (Rem.) obtained by EASIER for each iteration are displayed in Table~\ref{tab:details_CIFAR-10} for CIFAR-10, in Table~\ref{tab:details_Tiny} for Tiny-ImageNet-200, in Table~\ref{tab:details_PACS} for PACS, in Table~\ref{tab:details_VLCS} for VLCS, in Table~\ref{tab:details_Flowers} for Flowers-102, in Table~\ref{tab:details_DTD} for DTD, in Table~\ref{tab:details_Aircraft} for Aircraft.

\begin{table}[!ht]
    \centering
    \begin{tabular}{ccc|cc|cc|cc}
    \toprule
        ~ & \multicolumn{2}{c}{\textbf{~~ResNet-18~~}} & \multicolumn{2}{c}{\textbf{~~Swin-T~~}} & \multicolumn{2}{c}{\textbf{MobileNetv2}} & \multicolumn{2}{c}{\textbf{~~VGG-16~~}} \\ 
        ~ & ~~top-1~~ & ~~Rem.~~ & ~~top-1~~ & ~~Rem. & ~~top-1~~ & ~~Rem.~~ & ~~top-1 & ~~Rem.~~ \\
    \midrule
    \parbox[t]{5mm}{\multirow{35}{*}{\rotatebox[origin=c]{90}{EASIER on CIFAR-10}}} & 92,47 & 0/17 & 91,66 & 0/12 & 93,65 & 0/35 & 93,50 & 0/15 \\ 
        ~ & 92,39 & 1/17 & 91,95 & 1/12 & 93,60 & 1/35 & 93,54 & 1/15 \\ 
        ~ & 92,82 & 2/17 & 91,84 & 2/12 & 93,76 & 2/35 & 93,44 & 2/15 \\ 
        ~ & 92,15 & 3/17 & 91,57 & 3/12 & 93,67 & 3/35 & 93,58 & 3/15 \\ 
        ~ & 92,47 & 4/17 & 91,73 & 4/12 & 93,26 & 4/35 & 93,51 & 4/15 \\ 
        ~ & 91,97 & 5/17 & 91,40 & 5/12 & 93,14 & 5/35 & 93,62 & 5/15 \\ 
        ~ & 92,22 & 6/17 & 91,25 & 6/12 & 93,22 & 6/35 & 93,42 & 6/15 \\ 
        ~ & 92,40 & 7/17 & 91,41 & 7/12 & 92,99 & 7/35 & 93,12 & 7/15 \\ 
        ~ & 92,10 & 8/17 & 90,83 & 8/12 & 93,16 & 8/35 & 93,61 & 8/15 \\ 
        ~ & 90,35 & 9/17 & 90,12 & 9/12 & 92,91 & 9/35 & 92,65 & 9/15 \\ 
        ~ & 89,94 & 10/17 & 87,79 & 10/12 & 92,79 & 10/35 & 93,04 & 10/15 \\ 
        ~ & 86,53 & 11/17 & 87,96 & 11/12 & 93,03 & 11/35 & 92,44 & 11/15 \\ 
        ~ & 84,52 & 12/17 & ~ & ~ & 93,16 & 12/35 & 91,46 & 12/15 \\ 
        ~ & 80,69 & 13/17 & ~ & ~ & 92,79 & 13/35 & 90,65 & 13/15 \\ 
        ~ & 80,95 & 14/17 & ~ & ~ & 92,79 & 14/35 & 90,38 & 14/15 \\ 
        ~ & 78,98 & 15/17 & ~ & ~ & 92,77 & 15/35 & ~ & ~ \\ 
        ~ & 77,74 & 16/17 & ~ & ~ & 92,45 & 16/35 & ~ & ~ \\ 
        ~ & ~ & ~ & ~ & ~ & 92,05 & 17/35 & ~ & ~ \\ 
        ~ & ~ & ~ & ~ & ~ & 91,80 & 18/35 & ~ & ~ \\ 
        ~ & ~ & ~ & ~ & ~ & 90,89 & 19/35 & ~ & ~ \\
        ~ & ~ & ~ & ~ & ~ & 90,79 & 20/35 & ~ & ~ \\ 
        ~ & ~ & ~ & ~ & ~ & 90,84 & 21/35 & ~ & ~ \\ 
        ~ & ~ & ~ & ~ & ~ & 89,92 & 22/35 & ~ & ~ \\ 
        ~ & ~ & ~ & ~ & ~ & 90,33 & 23/35 & ~ & ~ \\ 
        ~ & ~ & ~ & ~ & ~ & 90,44 & 24/35 & ~ & ~ \\ 
        ~ & ~ & ~ & ~ & ~ & 89,42 & 25/35 & ~ & ~ \\ 
        ~ & ~ & ~ & ~ & ~ & 88,98 & 26/35 & ~ & ~ \\ 
        ~ & ~ & ~ & ~ & ~ & 89,56 & 27/35 & ~ & ~ \\ 
        ~ & ~ & ~ & ~ & ~ & 88,74 & 28/35 & ~ & ~ \\ 
        ~ & ~ & ~ & ~ & ~ & 88,43 & 29/35 & ~ & ~ \\ 
        ~ & ~ & ~ & ~ & ~ & 87,19 & 30/35 & ~ & ~ \\ 
        ~ & ~ & ~ & ~ & ~ & 87,67 & 31/35 & ~ & ~ \\ 
        ~ & ~ & ~ & ~ & ~ & 86,88 & 32/35 & ~ & ~ \\ 
        ~ & ~ & ~ & ~ & ~ & 84,97 & 33/35 & ~ & ~ \\ 
        ~ & ~ & ~ & ~ & ~ & 62,56 & 34/35 & ~ & ~ \\ 
    \bottomrule
    \\
    \end{tabular}
    \caption{Test performance (top-1) and the number of removed layers (Rem.) of EASIER on CIFAR-10.}
    \label{tab:details_CIFAR-10}
\end{table}

\begin{table}[!ht]
    \centering
    \begin{tabular}{ccc|cc|cc|cc}
    \toprule
        ~ & \multicolumn{2}{c}{\textbf{ResNet-18}} & \multicolumn{2}{c}{\textbf{Swin-T}} & \multicolumn{2}{c}{\textbf{MobileNetv2}} & \multicolumn{2}{c}{\textbf{VGG-16}} \\ 
        ~ & top-1 & Rem. & top-1 & Rem. & top-1 & Rem. & top-1 & Rem. \\
    \midrule
    \parbox[t]{5mm}{\multirow{35}{*}{\rotatebox[origin=c]{90}{EASIER on Tiny-ImageNet-200}}} & 41,26 & 0/17 & 75,78 & 0/12 & 46,54 & 0/35 & 63,94 & 0/15 \\ 
        ~ & 41,10 & 1/17 & 70,94 & 1/12 & 47,52 & 1/35 & 63,32 & 1/15 \\ 
        ~ &40,62 & 2/17 & 69,62 & 2/12 & 46,62 & 2/35 & 62,12 & 2/15 \\ 
        ~ &41,46 & 3/17 & 68,46 & 3/12 & 48,22 & 3/35 & 61,34 & 3/15 \\ 
        ~ &40,42 & 4/17 & 67,06 & 4/12 & 47,36 & 4/35 & 60,44 & 4/15 \\ 
        ~ &40,42 & 5/17 & 66,68 & 5/12 & 47,52 & 5/35 & 59,06 & 5/15 \\ 
        ~ &35,84 & 6/17 & 65,80 & 6/12 & 47,88 & 6/35 & 58,60 & 6/15 \\ 
        ~ &36,30 & 7/17 & 63,90 & 7/12 & 48,44 & 7/35 & 57,60 & 7/15 \\ 
        ~ &33,82 & 8/17 & 61,10 & 8/12 & 48,04 & 8/35 & 56,80 & 8/15 \\ 
        ~ &34,04 & 9/17 & 56,76 & 9/12 & 47,76 & 9/35 & 56,04 & 9/15 \\ 
        ~ &33,02 & 10/17 & 56,30 & 10/12 & 47,96 & 10/35 & 55,40 & 10/15 \\ 
        ~ &31,58 & 11/17 & 52,26 & 11/12 & 47,58 & 11/35 & 53,18 & 11/15 \\ 
        ~ &30,92 & 12/17 & ~ & ~ & 47,78 & 12/35 & 51,24 & 12/15 \\ 
        ~ &33,88 & 13/17 & ~ & ~ & 48,00 & 13/35 & 48,02 & 13/15 \\ 
        ~ &33,62 & 14/17 & ~ & ~ & 48,96 & 14/35 & 43,78 & 14/15 \\ 
        ~ &33,90 & 15/17 & ~ & ~ & 48,88 & 15/35 & ~ & ~ \\ 
        ~ &31,12 & 16/17 & ~ & ~ & 48,80 & 16/35 & ~ & ~ \\ 
        ~ & ~ & ~ & ~ & ~ & 48,54 & 17/35 & ~ & ~ \\ 
        ~ & ~ & ~ & ~ & ~ & 49,72 & 18/35 & ~ & ~ \\ 
        ~ & ~ & ~ & ~ & ~ & 49,50 & 19/35 & ~ & ~ \\ 
        ~ & ~ & ~ & ~ & ~ & 49,84 & 20/35 & ~ & ~ \\ 
        ~ & ~ & ~ & ~ & ~ & 50,14 & 21/35 & ~ & ~ \\ 
        ~ & ~ & ~ & ~ & ~ & 49,18 & 22/35 & ~ & ~ \\ 
        ~ & ~ & ~ & ~ & ~ & 49,88 & 23/35 & ~ & ~ \\ 
        ~ & ~ & ~ & ~ & ~ & 49,88 & 24/35 & ~ & ~ \\ 
        ~ & ~ & ~ & ~ & ~ & 49,92 & 25/35 & ~ & ~ \\ 
        ~ & ~ & ~ & ~ & ~ & 48,98 & 26/35 & ~ & ~ \\ 
        ~ & ~ & ~ & ~ & ~ & 49,16 & 27/35 & ~ & ~ \\ 
        ~ & ~ & ~ & ~ & ~ & 48,80 & 28/35 & ~ & ~ \\ 
        ~ & ~ & ~ & ~ & ~ & 45,70 & 29/35 & ~ & ~ \\ 
        ~ & ~ & ~ & ~ & ~ & 45,44 & 30/35 & ~ & ~ \\ 
        ~ & ~ & ~ & ~ & ~ & 45,30 & 31/35 & ~ & ~ \\ 
        ~ & ~ & ~ & ~ & ~ & 41,20 & 32/35 & ~ & ~ \\ 
        ~ & ~ & ~ & ~ & ~ & 35,50 & 33/35 & ~ & ~ \\ 
        ~ & ~ & ~ & ~ & ~ & 18,16 & 34/35 & ~ & ~ \\ 
    \bottomrule
    \\
    \end{tabular}
    \caption{Test performance (top-1) and the number of removed layers (Rem.) of EASIER on Tiny-ImageNet-200.}
    \label{tab:details_Tiny}
\end{table}

\begin{table}[!ht]
    \centering
    \begin{tabular}{ccc|cc|cc|cc}
    \toprule
        ~ & \multicolumn{2}{c}{\textbf{~~ResNet-18~~}} & \multicolumn{2}{c}{\textbf{~~Swin-T~~}} & \multicolumn{2}{c}{\textbf{MobileNetv2}} & \multicolumn{2}{c}{\textbf{~~VGG-16~~}} \\ 
        ~ & ~~top-1~~ & ~~Rem.~~ & ~~top-1~~ & ~~Rem. & ~~top-1~~ & ~~Rem.~~ & ~~top-1 & ~~Rem.~~ \\
    \midrule
    \parbox[t]{5mm}{\multirow{35}{*}{\rotatebox[origin=c]{90}{EASIER on PACS}}} & 79,70 & 0/17 & 97,10 & 0/12 & 95,60 & 0/35 & 95,40 & 0/15 \\ 
        ~ & 86,60 & 1/17 & 95,90 & 1/12 & 95,30 & 1/35 & 95,70 & 1/15 \\ 
        ~ & 86,40 & 2/17 & 95,90 & 2/12 & 95,40 & 2/35 & 95,20 & 2/15 \\ 
        ~ & 88,30 & 3/17 & 93,80 & 3/12 & 94,80 & 3/35 & 95,20 & 3/15 \\ 
        ~ & 89,00 & 4/17 & 94,30 & 4/12 & 95,20 & 4/35 & 95,50 & 4/15 \\ 
        ~ & 87,90 & 5/17 & 94,60 & 5/12 & 94,50 & 5/35 & 94,20 & 5/15 \\ 
        ~ & 87,70 & 6/17 & 92,50 & 6/12 & 94,50 & 6/35 & 93,00 & 6/15 \\ 
        ~ & 88,30 & 7/17 & 91,80 & 7/12 & 94,40 & 7/35 & 93,00 & 7/15 \\ 
        ~ & 87,30 & 8/17 & 91,50 & 8/12 & 94,20 & 8/35 & 91,70 & 8/15 \\ 
        ~ & 88,30 & 9/17 & 84,80 & 9/12 & 93,70 & 9/35 & 90,20 & 9/15 \\ 
        ~ & 87,70 & 10/17 & 85,90 & 10/12 & 93,80 & 10/35 & 91,40 & 10/15 \\ 
        ~ & 86,80 & 11/17 & 86,20 & 11/12 & 93,60 & 11/35 & 89,90 & 11/15 \\ 
        ~ & 87,10 & 12/17 & ~ & ~ & 92,70 & 12/35 & 87,40 & 12/15 \\ 
        ~ & 84,30 & 13/17 & ~ & ~ & 92,10 & 13/35 & 85,90 & 13/15 \\ 
        ~ & 80,90 & 14/17 & ~ & ~ & 92,20 & 14/35 & 85,50 & 14/15 \\ 
        ~ & 75,10 & 15/17 & ~ & ~ & 92,40 & 15/35 & ~ & ~ \\ 
        ~ & 52,00 & 16/17 & ~ & ~ & 88,60 & 16/35 & ~ & ~ \\ 
        ~ & ~ & ~ & ~ & ~ & 89,80 & 17/35 & ~ & ~ \\ 
        ~ & ~ & ~ & ~ & ~ & 89,60 & 18/35 & ~ & ~ \\ 
        ~ & ~ & ~ & ~ & ~ & 89,00 & 19/35 & ~ & ~ \\ 
        ~ & ~ & ~ & ~ & ~ & 91,00 & 20/35 & ~ & ~ \\ 
        ~ & ~ & ~ & ~ & ~ & 89,10 & 21/35 & ~ & ~ \\ 
        ~ & ~ & ~ & ~ & ~ & 88,30 & 22/35 & ~ & ~ \\ 
        ~ & ~ & ~ & ~ & ~ & 88,40 & 23/35 & ~ & ~ \\ 
        ~ & ~ & ~ & ~ & ~ & 87,90 & 24/35 & ~ & ~ \\ 
        ~ & ~ & ~ & ~ & ~ & 86,40 & 25/35 & ~ & ~ \\ 
        ~ & ~ & ~ & ~ & ~ & 86,30 & 26/35 & ~ & ~ \\ 
        ~ & ~ & ~ & ~ & ~ & 87,60 & 27/35 & ~ & ~ \\ 
        ~ & ~ & ~ & ~ & ~ & 87,30 & 28/35 & ~ & ~ \\ 
        ~ & ~ & ~ & ~ & ~ & 86,60 & 29/35 & ~ & ~ \\ 
        ~ & ~ & ~ & ~ & ~ & 79,50 & 30/35 & ~ & ~ \\ 
        ~ & ~ & ~ & ~ & ~ & 67,10 & 31/35 & ~ & ~ \\ 
        ~ & ~ & ~ & ~ & ~ & 63,30 & 32/35 & ~ & ~ \\ 
        ~ & ~ & ~ & ~ & ~ & 62,90 & 33/35 & ~ & ~ \\ 
        ~ & ~ & ~ & ~ & ~ & 43,80 & 34/35 & ~ & ~ \\ 
    \bottomrule
    \\
    \end{tabular}
    \caption{Test performance (top-1) and the number of removed layers (Rem.) of EASIER on PACS.}
    \label{tab:details_PACS}
\end{table}

\begin{table}[!ht]
    \centering
    \begin{tabular}{ccc|cc|cc|cc}
    \toprule
        ~ & \multicolumn{2}{c}{\textbf{~~ResNet-18~~}} & \multicolumn{2}{c}{\textbf{~~Swin-T~~}} & \multicolumn{2}{c}{\textbf{MobileNetv2}} & \multicolumn{2}{c}{\textbf{~~VGG-16~~}} \\ 
        ~ & ~~top-1~~ & ~~Rem.~~ & ~~top-1~~ & ~~Rem. & ~~top-1~~ & ~~Rem.~~ & ~~top-1 & ~~Rem.~~ \\
    \midrule
    \parbox[t]{5mm}{\multirow{35}{*}{\rotatebox[origin=c]{90}{EASIER on VLCS}}} & 68,13 & 0/17 & 83,04 & 0/12 & 81,36 & 0/35 & 82,76 & 0/15 \\ 
        ~ & 70,83 & 1/17 & 82,01 & 1/12 & 79,50 & 1/35 & 81,64 & 1/15 \\ 
        ~ & 72,04 & 2/17 & 81,45 & 2/12 & 79,59 & 2/35 & 81,92 & 2/15 \\ 
        ~ & 71,48 & 3/17 & 79,87 & 3/12 & 77,82 & 3/35 & 79,59 & 3/15 \\ 
        ~ & 71,39 & 4/17 & 78,75 & 4/12 & 78,56 & 4/35 & 80,99 & 4/15 \\ 
        ~ & 71,67 & 5/17 & 78,19 & 5/12 & 78,10 & 5/35 & 78,94 & 5/15 \\ 
        ~ & 71,20 & 6/17 & 79,12 & 6/12 & 77,17 & 6/35 & 78,84 & 6/15 \\ 
        ~ & 71,11 & 7/17 & 76,51 & 7/12 & 76,79 & 7/35 & 78,66 & 7/15 \\ 
        ~ & 70,92 & 8/17 & 78,19 & 8/12 & 77,82 & 8/35 & 78,29 & 8/15 \\ 
        ~ & 71,85 & 9/17 & 78,19 & 9/12 & 78,19 & 9/35 & 78,56 & 9/15 \\ 
        ~ & 71,30 & 10/17 & 75,68 & 10/12 & 76,14 & 10/35 & 76,51 & 10/15 \\ 
        ~ & 71,02 & 11/17 & 73,63 & 11/12 & 76,61 & 11/35 & 75,77 & 11/15 \\ 
        ~ & 70,83 & 12/17 & ~ & ~ & 76,23 & 12/35 & 75,49 & 12/15 \\ 
        ~ & 69,80 & 13/17 & ~ & ~ & 78,56 & 13/35 & 76,05 & 13/15 \\ 
        ~ & 70,27 & 14/17 & ~ & ~ & 76,51 & 14/35 & 70,18 & 14/15 \\ 
        ~ & 54,24 & 15/17 & ~ & ~ & 76,79 & 15/35 & ~ & ~ \\ 
        ~ & 55,27 & 16/17 & ~ & ~ & 74,84 & 16/35 & ~ & ~ \\ 
        ~ & ~ & ~ & ~ & ~ & 74,18 & 17/35 & ~ & ~ \\ 
        ~ & ~ & ~ & ~ & ~ & 74,74 & 18/35 & ~ & ~ \\ 
        ~ & ~ & ~ & ~ & ~ & 73,44 & 19/35 & ~ & ~ \\ 
        ~ & ~ & ~ & ~ & ~ & 75,58 & 20/35 & ~ & ~ \\ 
        ~ & ~ & ~ & ~ & ~ & 73,90 & 21/35 & ~ & ~ \\ 
        ~ & ~ & ~ & ~ & ~ & 72,88 & 22/35 & ~ & ~ \\ 
        ~ & ~ & ~ & ~ & ~ & 73,07 & 23/35 & ~ & ~ \\ 
        ~ & ~ & ~ & ~ & ~ & 74,74 & 24/35 & ~ & ~ \\ 
        ~ & ~ & ~ & ~ & ~ & 72,69 & 25/35 & ~ & ~ \\ 
        ~ & ~ & ~ & ~ & ~ & 73,07 & 26/35 & ~ & ~ \\ 
        ~ & ~ & ~ & ~ & ~ & 72,88 & 27/35 & ~ & ~ \\ 
        ~ & ~ & ~ & ~ & ~ & 71,48 & 28/35 & ~ & ~ \\ 
        ~ & ~ & ~ & ~ & ~ & 72,88 & 29/35 & ~ & ~ \\ 
        ~ & ~ & ~ & ~ & ~ & 71,85 & 30/35 & ~ & ~ \\ 
        ~ & ~ & ~ & ~ & ~ & 71,02 & 31/35 & ~ & ~ \\ 
        ~ & ~ & ~ & ~ & ~ & 71,76 & 32/35 & ~ & ~ \\ 
        ~ & ~ & ~ & ~ & ~ & 64,96 & 33/35 & ~ & ~ \\ 
        ~ & ~ & ~ & ~ & ~ & 52,47 & 34/35 & ~ & ~ \\ 
    \bottomrule
    \\
    \end{tabular}
    \caption{Test performance (top-1) and the number of removed layers (Rem.) of EASIER on VLCS.}
    \label{tab:details_VLCS}
\end{table}

\begin{table}[!ht]
    \centering
    \begin{tabular}{ccc|cc|cc|cc}
    \toprule
        ~ & \multicolumn{2}{c}{\textbf{~~ResNet-18~~}} & \multicolumn{2}{c}{\textbf{~~Swin-T~~}} & \multicolumn{2}{c}{\textbf{MobileNetv2}} & \multicolumn{2}{c}{\textbf{~~VGG-16~~}} \\ 
        ~ & ~~top-1~~ & ~~Rem.~~ & ~~top-1~~ & ~~Rem. & ~~top-1~~ & ~~Rem.~~ & ~~top-1 & ~~Rem.~~ \\
    \midrule
    \parbox[t]{5mm}{\multirow{35}{*}{\rotatebox[origin=c]{90}{EASIER on Flowers-102}}} & 88,88 & 0/17 & 92,70 & 0/12 & 88,50 & 0/35 & 86,47 & 0/15 \\ 
        ~ & 88,23 & 1/17 & 93,10 & 1/12 & 89,56 & 1/35 & 86,32 & 1/15 \\ 
        ~ & 87,49 & 2/17 & 92,71 & 2/12 & 89,79 & 2/35 & 88,81 & 2/15 \\ 
        ~ & 87,53 & 3/17 & 88,93 & 3/12 & 90,31 & 3/35 & 88,32 & 3/15 \\ 
        ~ & 87,25 & 4/17 & 90,93 & 4/12 & 89,79 & 4/35 & 84,63 & 4/15 \\ 
        ~ & 84,57 & 5/17 & 88,89 & 5/12 & 89,82 & 5/35 & 85,69 & 5/15 \\ 
        ~ & 83,43 & 6/17 & 87,49 & 6/12 & 89,27 & 6/35 & 84,14 & 6/15 \\ 
        ~ & 63,41 & 7/17 & 85,92 & 7/12 & 89,02 & 7/35 & 82,60 & 7/15 \\ 
        ~ & 73,12 & 8/17 & 85,10 & 8/12 & 87,93 & 8/35 & 84,62 & 8/15 \\ 
        ~ & 64,58 & 9/17 & 80,39 & 9/12 & 88,49 & 9/35 & 81,53 & 9/15 \\ 
        ~ & 70,81 & 10/17 & 80,73 & 10/12 & 88,37 & 10/35 & 80,68 & 10/15 \\ 
        ~ & 72,22 & 11/17 & 82,05 & 11/12 & 87,02 & 11/35 & 77,46 & 11/15 \\ 
        ~ & 67,28 & 12/17 & ~ & ~ & 87,56 & 12/35 & 75,15 & 12/15 \\ 
        ~ & 65,33 & 13/17 & ~ & ~ & 86,60 & 13/35 & 74,84 & 13/15 \\ 
        ~ & 28,17 & 14/17 & ~ & ~ & 87,01 & 14/35 & 69,47 & 14/15 \\ 
        ~ & 32,85 & 15/17 & ~ & ~ & 86,26 & 15/35 & ~ & ~ \\ 
        ~ & 20,15 & 16/17 & ~ & ~ & 86,06 & 16/35 & ~ & ~ \\ 
        ~ & ~ & ~ & ~ & ~ & 83,56 & 17/35 & ~ & ~ \\ 
        ~ & ~ & ~ & ~ & ~ & 82,39 & 18/35 & ~ & ~ \\ 
        ~ & ~ & ~ & ~ & ~ & 82,83 & 19/35 & ~ & ~ \\ 
        ~ & ~ & ~ & ~ & ~ & 83,61 & 20/35 & ~ & ~ \\ 
        ~ & ~ & ~ & ~ & ~ & 82,92 & 21/35 & ~ & ~ \\ 
        ~ & ~ & ~ & ~ & ~ & 82,40 & 22/35 & ~ & ~ \\ 
        ~ & ~ & ~ & ~ & ~ & 82,92 & 23/35 & ~ & ~ \\ 
        ~ & ~ & ~ & ~ & ~ & 81,36 & 24/35 & ~ & ~ \\ 
        ~ & ~ & ~ & ~ & ~ & 81,14 & 25/35 & ~ & ~ \\ 
        ~ & ~ & ~ & ~ & ~ & 74,63 & 26/35 & ~ & ~ \\ 
        ~ & ~ & ~ & ~ & ~ & 68,30 & 27/35 & ~ & ~ \\ 
        ~ & ~ & ~ & ~ & ~ & 69,62 & 28/35 & ~ & ~ \\ 
        ~ & ~ & ~ & ~ & ~ & 68,08 & 29/35 & ~ & ~ \\ 
        ~ & ~ & ~ & ~ & ~ & 57,78 & 30/35 & ~ & ~ \\ 
        ~ & ~ & ~ & ~ & ~ & 58,74 & 31/35 & ~ & ~ \\ 
        ~ & ~ & ~ & ~ & ~ & 43,10 & 32/35 & ~ & ~ \\ 
        ~ & ~ & ~ & ~ & ~ & 40,97 & 33/35 & ~ & ~ \\ 
        ~ & ~ & ~ & ~ & ~ & 20,28 & 34/35 & ~ & ~ \\ 
    \bottomrule
    \\
    \end{tabular}
    \caption{Test performance (top-1) and the number of removed layers (Rem.) of EASIER on Flowers-102.}
    \label{tab:details_Flowers}
\end{table}

\begin{table}[!ht]
    \centering
    \begin{tabular}{ccc|cc|cc|cc}
    \toprule
        ~ & \multicolumn{2}{c}{\textbf{~~ResNet-18~~}} & \multicolumn{2}{c}{\textbf{~~Swin-T~~}} & \multicolumn{2}{c}{\textbf{MobileNetv2}} & \multicolumn{2}{c}{\textbf{~~VGG-16~~}} \\ 
        ~ & ~~top-1~~ & ~~Rem.~~ & ~~top-1~~ & ~~Rem. & ~~top-1~~ & ~~Rem.~~ & ~~top-1 & ~~Rem.~~ \\
    \midrule
    \parbox[t]{5mm}{\multirow{35}{*}{\rotatebox[origin=c]{90}{EASIER on DTD}}} & 60,53 & 0/17 & 67,50 & 0/12 & 64,41 & 0/35 & 64,20 & 0/15 \\ 
        ~ & 61,97 & 1/17 & 70,05 & 1/12 & 63,09 & 1/35 & 64,57 & 1/15 \\ 
        ~ & 61,22 & 2/17 & 67,02 & 2/12 & 62,93 & 2/35 & 64,73 & 2/15 \\ 
        ~ & 62,02 & 3/17 & 66,54 & 3/12 & 62,39 & 3/35 & 64,57 & 3/15 \\ 
        ~ & 59,73 & 4/17 & 63,67 & 4/12 & 62,82 & 4/35 & 63,62 & 4/15 \\ 
        ~ & 58,99 & 5/17 & 62,23 & 5/12 & 63,67 & 5/35 & 59,84 & 5/15 \\ 
        ~ & 57,55 & 6/17 & 58,99 & 6/12 & 63,83 & 6/35 & 59,95 & 6/15 \\ 
        ~ & 53,40 & 7/17 & 58,94 & 7/12 & 62,45 & 7/35 & 58,30 & 7/15 \\ 
        ~ & 49,26 & 8/17 & 58,19 & 8/12 & 61,17 & 8/35 & 58,19 & 8/15 \\ 
        ~ & 52,29 & 9/17 & 57,18 & 9/12 & 62,29 & 9/35 & 56,01 & 9/15 \\ 
        ~ & 50,48 & 10/17 & 52,45 & 10/12 & 60,11 & 10/35 & 54,10 & 10/15 \\ 
        ~ & 52,61 & 11/17 & 53,19 & 11/12 & 61,97 & 11/35 & 49,41 & 11/15 \\ 
        ~ & 49,68 & 12/17 & ~ & ~ & 60,80 & 12/35 & 47,29 & 12/15 \\ 
        ~ & 41,86 & 13/17 & ~ & ~ & 60,74 & 13/35 & 47,34 & 13/15 \\ 
        ~ & 32,55 & 14/17 & ~ & ~ & 60,59 & 14/35 & 41,54 & 14/15 \\ 
        ~ & 33,35 & 15/17 & ~ & ~ & 61,01 & 15/35 & ~ & ~ \\ 
        ~ & 18,94 & 16/17 & ~ & ~ & 61,54 & 16/35 & ~ & ~ \\ 
        ~ & ~ & ~ & ~ & ~ & 61,60 & 17/35 & ~ & ~ \\ 
        ~ & ~ & ~ & ~ & ~ & 59,15 & 18/35 & ~ & ~ \\ 
        ~ & ~ & ~ & ~ & ~ & 60,96 & 19/35 & ~ & ~ \\ 
        ~ & ~ & ~ & ~ & ~ & 59,89 & 20/35 & ~ & ~ \\ 
        ~ & ~ & ~ & ~ & ~ & 57,02 & 21/35 & ~ & ~ \\ 
        ~ & ~ & ~ & ~ & ~ & 57,07 & 22/35 & ~ & ~ \\ 
        ~ & ~ & ~ & ~ & ~ & 52,07 & 23/35 & ~ & ~ \\ 
        ~ & ~ & ~ & ~ & ~ & 53,83 & 24/35 & ~ & ~ \\ 
        ~ & ~ & ~ & ~ & ~ & 55,64 & 25/35 & ~ & ~ \\ 
        ~ & ~ & ~ & ~ & ~ & 54,20 & 26/35 & ~ & ~ \\ 
        ~ & ~ & ~ & ~ & ~ & 43,88 & 27/35 & ~ & ~ \\ 
        ~ & ~ & ~ & ~ & ~ & 43,67 & 28/35 & ~ & ~ \\ 
        ~ & ~ & ~ & ~ & ~ & 43,83 & 29/35 & ~ & ~ \\ 
        ~ & ~ & ~ & ~ & ~ & 32,18 & 30/35 & ~ & ~ \\ 
        ~ & ~ & ~ & ~ & ~ & 32,50 & 31/35 & ~ & ~ \\ 
        ~ & ~ & ~ & ~ & ~ & 30,90 & 32/35 & ~ & ~ \\ 
        ~ & ~ & ~ & ~ & ~ & 28,62 & 33/35 & ~ & ~ \\ 
        ~ & ~ & ~ & ~ & ~ & 14,89 & 34/35 & ~ & ~ \\ 
    \bottomrule
    \\
    \end{tabular}
    \caption{Test performance (top-1) and the number of removed layers (Rem.) of EASIER on DTD.}
    \label{tab:details_DTD}
\end{table}

\begin{table}[!ht]
    \centering
    \begin{tabular}{ccc|cc|cc|cc}
    \toprule
        ~ & \multicolumn{2}{c}{\textbf{~~ResNet-18~~}} & \multicolumn{2}{c}{\textbf{~~Swin-T~~}} & \multicolumn{2}{c}{\textbf{MobileNetv2}} & \multicolumn{2}{c}{\textbf{~~VGG-16~~}} \\ 
        ~ & ~~top-1~~ & ~~Rem.~~ & ~~top-1~~ & ~~Rem. & ~~top-1~~ & ~~Rem.~~ & ~~top-1 & ~~Rem.~~ \\
    \midrule
    \parbox[t]{5mm}{\multirow{35}{*}{\rotatebox[origin=c]{90}{EASIER on Aircraft}}} & 73,36 & 0/17 & 76,39 & 0/12 & 73,36 & 0/35 & 75,85 & 0/15 \\ 
        ~ & 72,79 & 1/17 & 76,81 & 1/12 & 73,00 & 1/35 & 78,13 & 1/15 \\ 
        ~ & 71,59 & 2/17 & 75,97 & 2/12 & 73,69 & 2/35 & 76,69 & 2/15 \\ 
        ~ & 71,14 & 3/17 & 74,65 & 3/12 & 72,13 & 3/35 & 76,63 & 3/15 \\ 
        ~ & 70,03 & 4/17 & 76,15 & 4/12 & 72,55 & 4/35 & 75,70 & 4/15 \\ 
        ~ & 71,17 & 5/17 & 74,59 & 5/12 & 71,74 & 5/35 & 66,52 & 5/15 \\ 
        ~ & 71,14 & 6/17 & 74,74 & 6/12 & 71,20 & 6/35 & 69,70 & 6/15 \\ 
        ~ & 65,47 & 7/17 & 74,44 & 7/12 & 70,30 & 7/35 & 67,21 & 7/15 \\ 
        ~ & 61,36 & 8/17 & 72,13 & 8/12 & 68,98 & 8/35 & 65,32 & 8/15 \\ 
        ~ & 60,85 & 9/17 & 71,62 & 9/12 & 69,31 & 9/35 & 64,12 & 9/15 \\ 
        ~ & 63,19 & 10/17 & 69,76 & 10/12 & 68,92 & 10/35 & 56,83 & 10/15 \\ 
        ~ & 66,22 & 11/17 & 71,41 & 11/12 & 69,52 & 11/35 & 55,42 & 11/15 \\ 
        ~ & 62,89 & 12/17 & ~ & ~ & 69,16 & 12/35 & 57,82 & 12/15 \\ 
        ~ & 39,12 & 13/17 & ~ & ~ & 67,60 & 13/35 & 49,23 & 13/15 \\ 
        ~ & 35,85 & 14/17 & ~ & ~ & 67,60 & 14/35 & 52,57 & 14/15 \\ 
        ~ & 28,11 & 15/17 & ~ & ~ & 67,15 & 15/35 & ~ & ~ \\ 
        ~ & 15,63 & 16/17 & ~ & ~ & 63,97 & 16/35 & ~ & ~ \\ 
        ~ & ~ & ~ & ~ & ~ & 64,66 & 17/35 & ~ & ~ \\ 
        ~ & ~ & ~ & ~ & ~ & 63,46 & 18/35 & ~ & ~ \\ 
        ~ & ~ & ~ & ~ & ~ & 64,30 & 19/35 & ~ & ~ \\ 
        ~ & ~ & ~ & ~ & ~ & 62,26 & 20/35 & ~ & ~ \\ 
        ~ & ~ & ~ & ~ & ~ & 63,61 & 21/35 & ~ & ~ \\ 
        ~ & ~ & ~ & ~ & ~ & 62,68 & 22/35 & ~ & ~ \\ 
        ~ & ~ & ~ & ~ & ~ & 52,75 & 23/35 & ~ & ~ \\ 
        ~ & ~ & ~ & ~ & ~ & 55,57 & 24/35 & ~ & ~ \\ 
        ~ & ~ & ~ & ~ & ~ & 53,89 & 25/35 & ~ & ~ \\ 
        ~ & ~ & ~ & ~ & ~ & 51,67 & 26/35 & ~ & ~ \\ 
        ~ & ~ & ~ & ~ & ~ & 52,09 & 27/35 & ~ & ~ \\ 
        ~ & ~ & ~ & ~ & ~ & 51,85 & 28/35 & ~ & ~ \\ 
        ~ & ~ & ~ & ~ & ~ & 39,27 & 29/35 & ~ & ~ \\ 
        ~ & ~ & ~ & ~ & ~ & 40,08 & 30/35 & ~ & ~ \\ 
        ~ & ~ & ~ & ~ & ~ & 42,96 & 31/35 & ~ & ~ \\ 
        ~ & ~ & ~ & ~ & ~ & 22,17 & 32/35 & ~ & ~ \\ 
        ~ & ~ & ~ & ~ & ~ & 18,15 & 33/35 & ~ & ~ \\ 
        ~ & ~ & ~ & ~ & ~ & 6,69 & 34/35 & ~ & ~ \\ 
    \bottomrule
    \\
    \end{tabular}
    \caption{Test performance (top-1) and the number of removed layers (Rem.) of EASIER on Aircraft.}
    \label{tab:details_Aircraft}
\end{table}

\end{document}